\pdfoutput=1

\documentclass[11pt]{article}

\usepackage[final]{acl}

\usepackage{times}
\usepackage{latexsym}

\usepackage[T1]{fontenc}

\usepackage[utf8]{inputenc}

\usepackage{microtype}

\usepackage{inconsolata}

\usepackage{graphicx}
\usepackage{booktabs}
\usepackage{array}
\usepackage{tabularx}
\usepackage{amsmath}

\usepackage{lipsum,afterpage,refcount}
\newcommand{\setfootnotemark}{%
  \refstepcounter{footnote}%
  \footnotemark[\value{footnote}]}

\newenvironment{itemizesquish}{\begin{list}{\labelitemi}{\setlength{\itemsep}{-0.25em}\setlength{\labelwidth}{1em}\setlength{\leftmargin}{\labelwidth}\addtolength{\leftmargin}{\labelsep}}}{\end{list}}

\title{Latin Treebanks in Review: An Evaluation of Morphological Tagging Across Time}

\author{
    Marisa Hudspeth, Brendan O'Connor, Laure Thompson   \\
    University of Massachusetts, Amherst \\
    \texttt{\{mhudspeth, brenocon, laurejt\}@cs.umass.edu}
    }

\begin{document}
\maketitle
\begin{abstract}
Existing Latin treebanks draw from Latin's long written tradition, spanning 17 centuries and a variety of cultures.
Recent efforts have begun to harmonize these treebanks' annotations to better train and evaluate morphological taggers.
However, the heterogeneity of these treebanks must be carefully considered to build effective and reliable data.
In this work, we review existing Latin treebanks to identify the texts they draw from, identify their overlap, and document their coverage across time and genre.
We additionally design automated conversions of their morphological feature annotations into the conventions of standard Latin grammar.
From this, we build new time-period data splits that draw from the existing treebanks which we use to perform a broad cross-time analysis for POS and morphological feature tagging.
We find that BERT-based taggers outperform existing taggers while also being more robust to cross-domain shifts.

\end{abstract}

\section{Introduction}

Large-scale digitized Latin archives now document cultures across many centuries 
in wide a variety of genres from literature to legal documents.
With increasingly powerful Latin natural language processing tools \cite[e.g.][]{bamman2020latin,burns2023latincy},
morphological feature tagging is a promising method for Latin-based computational humanities.
The goal of morphological tagging is to identify a set of morphological feature-value pairs for each token of a given sentence.
These features can help researchers analyze agency, power, and other morphosyntactically-signalled phenomena which have been fruitfully investigated in English \cite{sap-etal-2017-connotation, greene-resnik-2009-words} and other languages \cite{rashkin-etal-2017-multilingual}.
For example, Voice (active, passive verbs) and Case (e.g., nominative, accusative ablative nouns) are useful for studying power and agency.

Although Latin taggers have relatively good performance, in our experience they often perform poorly on rarer feature values---such as passive voice--that may prove crucial for downstream analyses.
%
Toward this end, we hope to develop a Latin morphological tagger whose accuracy is robust across time and genre by leveraging the recent development of five separate Latin Universal Dependencies (UD; \citealp{de-marneffe-etal-2021-universal}) treebanks and recent efforts to harmonize their morphological tags \cite{gamba-zeman-2023-latin}. 
In this work we review these harmonized treebanks\footnote{Perseus \cite{perseus}, PROIEL \cite{proiel}, LLCT \cite{llct}, ITTB \cite{ittb}, and UDante \cite{udante}} plus the non-UD treebank LASLA \cite{denooz2004opera-lasla},
and conclude that more data curation is required to fully evaluate and improve morphological tagging's cross-domain accuracy.

\begin{figure}[t]
    \centering
    \includegraphics[width=\linewidth]{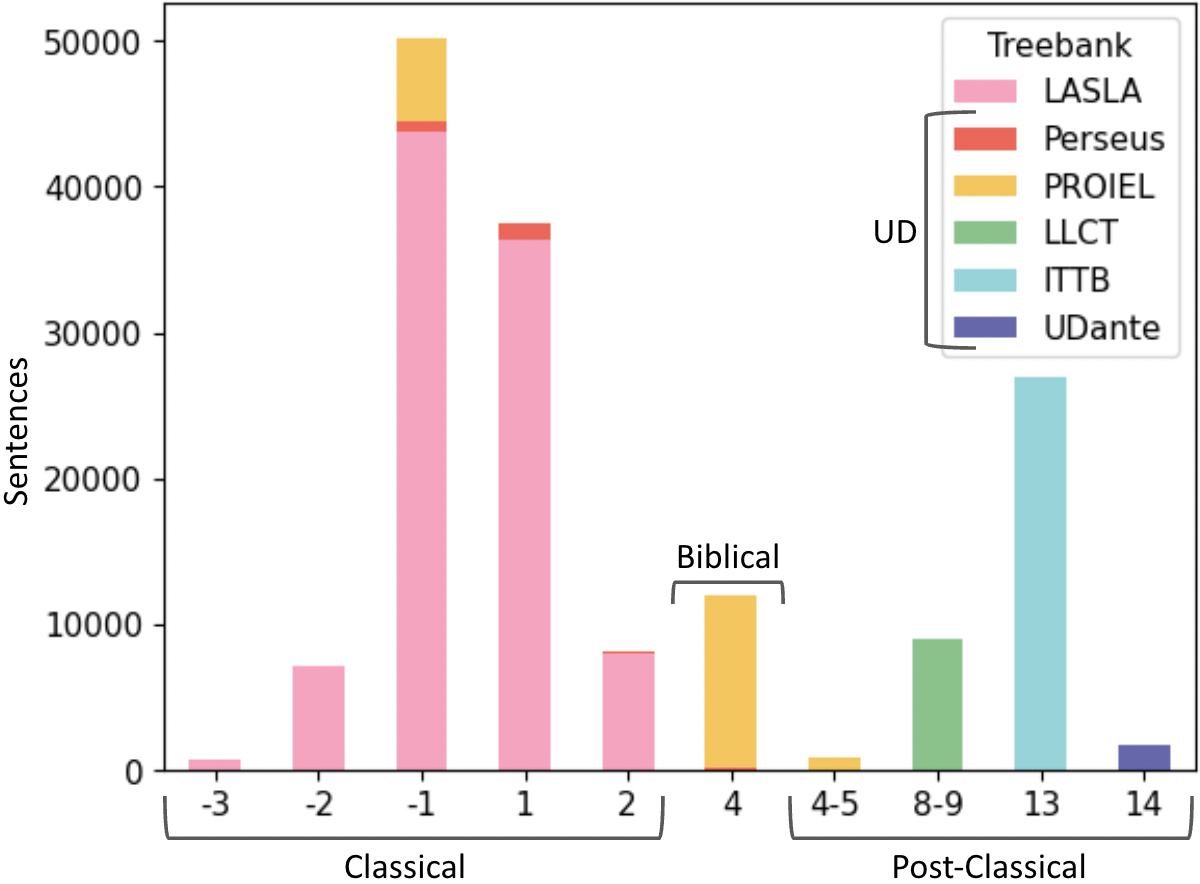}
    \caption{From our curated metadata (\S\ref{sec:data}), the number of sentences per century (3rd BCE---14th CE) across the 5 UD treebanks and LASLA, shown with three broad time periods.}
    \label{fig:timeline}
\end{figure}
Our contributions include: 
\textbf{1)} precisely documenting genre and historical context for the 544 texts 
within the UD treebanks as a machine-readable, cross-treebank resource that will enable future work to examine morphosyntactic association against these variables; 
\textbf{2)} harmonizing the UD and LASLA treebanks to reduce annotation differences that can affect training; 
\textbf{3)} proposing edits to the UD tagset that better align with standard analyses of Latin grammar 
to facilitate work by researchers with standard Latin training; and
\textbf{4)} conducting a cross-time analysis with experimental results broken down by historical period that show the promise of our harmonization efforts and BERT-based morphological taggers.\footnote{We have publicly released our new text-level metadata, standardized morphologically tagged text from described treebanks, and conversion software on \href{https://github.com/slanglab/latin-standardized-treebanks}{Github}.
} 


\section{Latin Treebanks Revisited}\label{sec:data}


\begin{table*}[]
    \centering
    \footnotesize
    \setlength{\tabcolsep}{2.5pt}
    \begin{tabular}{ll|l|l|l|l}
    \toprule
& Name & Text Data & Standard & Data & Paper/Version \\ 
               & & & Grammar? & Source & \\
\midrule
1 & Pre-UD & 4 non-UD & Mixed & \href{https://github.com/PerseusDL/treebank_data/tree/master}{Perseus}, \href{https://github.com/proiel/proiel-treebank}{PROIEL}   & \citealt{perseus, proiel}\\
& & & & \href{https://zenodo.org/records/3633607}{LLCT1},\href{https://itreebank.marginalia.it/view/download.php}{ITTB} &  \citealt{llct1, ittb} \\
2 & UD v2.8+ & 5 UD & Mixed &  \href{https://universaldependencies.org/\#language-la}{UD Site} & UD v2.8-11\\
3 & LatinCy Edits & 5 UD & Yes\setfootnotemark\label{latincy} & Unreleased & \citealt{burns2023latincy}\\ 
4 & Harmonized UD & 5 UD & No 
& \href{https://github.com/fjambe/Latin-variability}{Github} \textit{(acc.~1/24)} 
& UD v2.13; \citealt{gamba-zeman-2023-latin}\\
5 & LASLA & 1 non-UD & Mixed 
& \href{https://github.com/CIRCSE/LASLA/tree/main}{Github} \textit{(acc.~2/24)} 
& \citealt{denooz2004opera-lasla}\\ 
6 & EvaLatin 2022\setfootnotemark\label{evalatin_2022} & near-subset of 5\setfootnotemark\label{evalatin} & Mixed & \href{https://github.com/CIRCSE/LT4HALA/tree/master/2022/data_and_doc}{Github} & \citealt{sprugnoli-etal-2022-overview-evalatin}\\ 
7 & CIRCSE & 1 UD; subset of 5 & Mixed & \href{https://github.com/UniversalDependencies/UD_Latin-CIRCSE}{Github} & UD v2.14\\ 
\midrule 
8 & Harmonized + & 5 UD + LASLA, & Yes & \href{https://github.com/slanglab/latin-standardized-treebanks}{Github} & This work \\
  & Standardized & New Splits & &  & \\
         \bottomrule
    \end{tabular}
    \caption{Summary of data sources and history of Latin treebanks (for morphological tagging only).
    }
    \label{tab:datasets}
\end{table*}
\footnotetext[\getrefnumber{latincy}]{Although unreleased, we determined the feature-value set by examining LatinCy's outputs.}
\footnotetext[\getrefnumber{evalatin_2022}]{EvaLatin 2020 also has annotated data that is not directly sourced from LASLA but consists of a subset of LASLA's texts. This data is not annotated with morphological features.}
\footnotetext[\getrefnumber{evalatin}]{EvaLatin 2022 is a near-subset of LASLA because it has one non-Classical text that is not in LASLA.}

\subsection{Time and Genre Metadata} \label{sec:time_genre_metadata}
Detailed metadata on the texts included in the Latin UD treebanks is difficult to aggregate or lacking altogether. 
Information on the included works' time period, genre, author, and relative size has not been compiled in one place.
Our work takes major steps to fill this gap.
For all 544 texts across the five UD treebanks, we manually collected the following metadata: the source treebank, time period, century, internal treebank identifiers, cumulative and split-level sentence counts, author, and exhaustive genre labels.
\begin{figure}[h]
    \centering
    \includegraphics[width=\linewidth]{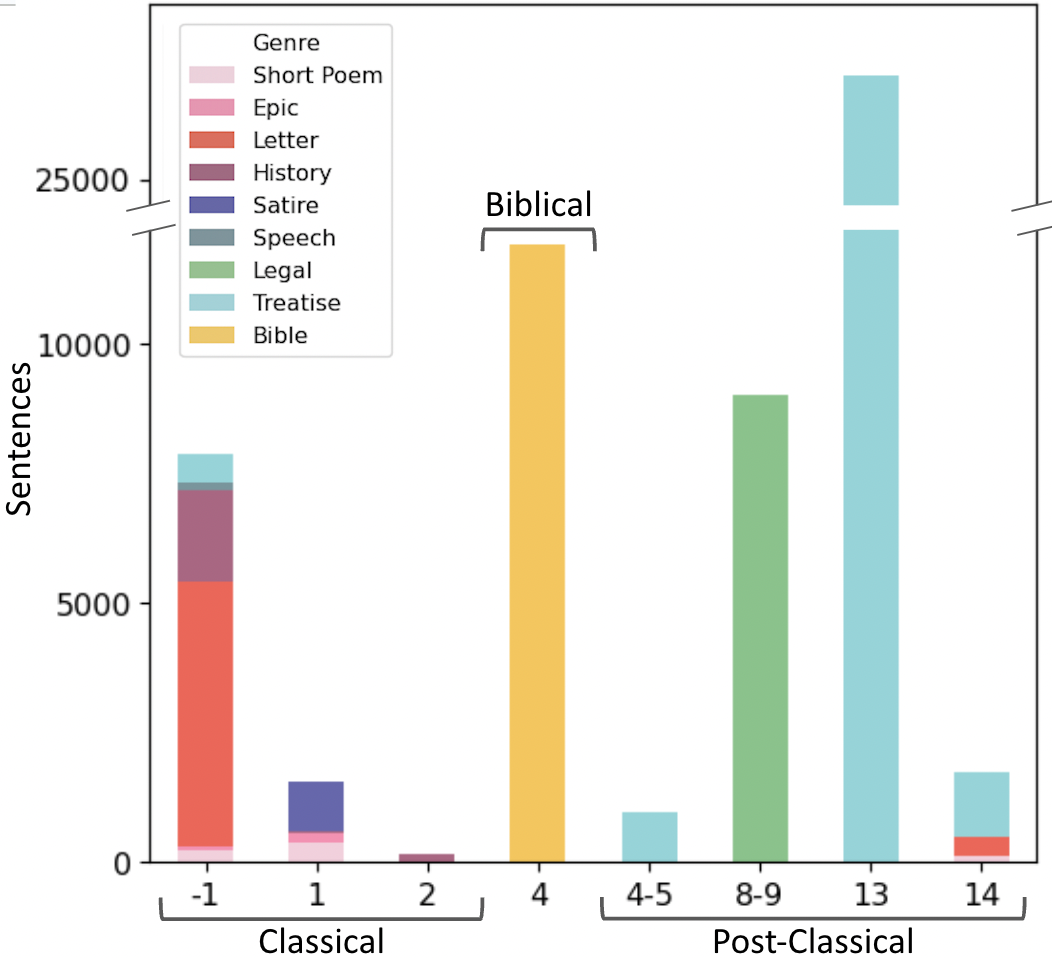}
    \caption{Number of sentences in the UD treebanks per century, colored by genre.}
    \label{fig:timeline_genre}
\end{figure}


\paragraph{Genre.} Figure \ref{fig:timeline_genre} shows the genre coverage of the UD treebanks. 
Previous EvaLatin campaigns \cite{sprugnoli-etal-2020-overview-evalatin, sprugnoli-etal-2022-overview-evalatin} have implicitly defined several genres (prose, poetry, epics, and histories), which were then used to analyze cross-genre tagging accuracy on Classical era, non-UD data. We expand upon these genres by including more fine-grained labels and by covering non-Classical texts.

We annotate nine exclusive genres: short poems, epics, letters, histories, satires, speeches, legal texts, treatises, and the Bible.\footnote{We also annotate additional non-exclusive genres (\S\ref{sec:ud_genres_appendix}).} 

\paragraph{Time.} We define the (approximate) century of each text
(Figure \ref{fig:timeline} and \ref{fig:timeline_genre}).  For cross-time analysis, we define three very broad time periods:
\vspace{-3mm}
\begin{itemizesquish}
\item \textbf{Classical} is defined as 3rd century BCE through the 2nd century CE, in line with conventional definitions of the Classical Latin language and periods \cite{brittanica-latin-language} and previous Latin NLP literature \cite{sprugnoli-etal-2020-overview-evalatin, sprugnoli-etal-2022-overview-evalatin}.

\item \textbf{Biblical} is defined as its own genre and time period, consisting solely of Jerome's \emph{Vulgata} from the 4th century CE.  It is significantly different from other texts given it is a translation (from much earlier material), and has relatively simpler grammar  \citep{Nunn}.

\item \textbf{Post-Classical} is defined as 4th century CE and later, excluding the Bible, thus including Late and Medieval Latin texts.  For simplicity, we do not split it further.
\end{itemizesquish}
\vspace{-3mm}

\noindent
Prior work in cross-time tagging either used a smaller set of time periods \cite{sprugnoli-etal-2022-overview-evalatin} or considered each UD treebank its own time period \cite{gamba-zeman-2023-latin}, which we argue is too approximate given our metadata findings (\S\ref{sec:ud_treebanks}).


\subsection{UD Treebanks} \label{sec:ud_treebanks}
Currently, there are five UD treebanks for Latin.\footnote{In May 2024, a sixth, CIRCSE, was added; it is a subset of LASLA.
} Four of these---Perseus, PROIEL, LLCT, and ITTB---were automatically converted to UD format, while the fifth, UDante, was annotated directly in UD.  Collectively, this corresponds to about 58,000 annotated sentences and 979,280 annotated tokens.
As Figure \ref{fig:timeline} shows, these treebanks cover a wide range of time but far from evenly.
We find that the three Post-Classical treebanks (LLCT, ITTB, and UDante) are quite distinct from each other in terms of genre and time period. LLCT consists entirely of medieval legal charters from the 8th and 9th centuries. ITTB consists of three philosophical and religious works by Thomas Aquinas from the 13th century. Finally, UDante is comprised of Dante Alighieri’s 14th century Latin works, including treatises, letters, and poems.

The two remaining treebanks, Perseus and PROIEL, are more diverse.
Most texts in Perseus are Classical, although 154 sentences are from Jerome’s \textit{Vulgata} (the Book of Revelation). While PROIEL also includes Classical texts, the majority (11785) of its 18411 sentences are also taken from Jerome’s \textit{Vulgata}. 
There is overlap between Perseus and PROIEL, as both share at least 145 sentences from the Book of Revelation.
\footnote{See Table \ref{tab:same_toks_pers_proiel_harmonized} for a breakdown of annotation agreement between these duplicate sentences.} Aside from Classical and Biblical texts, PROIEL also includes one 4th-5th century work, \textit{Opus Agriculturae} by Palladius.

\subsection{LASLA: Additional Classical-era treebank}
LASLA is a large, non-UD treebank for Latin \cite{denooz2004opera-lasla}.
By our own count, LASLA has 134 unique texts with 95,974 sentences and about 1.8M tokens.\footnote{A full list of authors, works, and tokens per text is available \hyperlink{https://www.lasla.uliege.be/cms/c_8570472/fr/lasla-textes-latins-traites-par-auteur}{here}.}
All texts are Classical. All genres included in UD are covered, in addition to plays.
Unlike the UD treebanks, LASLA does not have dependency relations.

\section{Harmonizing UD and LASLA Annotations} \label{sec:harmonizing_ud_lasla}

In this section, we describe steps taken to reduce the annotation differences between the Harmonized UD treebanks (Table \ref{tab:datasets} row 4) and LASLA (\ref{tab:datasets} row 5). Throughout this section, we sometimes use "UD" as a shorthand for \citet{gamba-zeman-2023-latin}'s Harmonized UD treebanks.

In \S\ref{sec:annot_agreement_before}, we outline the annotation agreement between Harmonized UD and LASLA before any intervention on our part. Then, we describe two types of changes: harmonization (\S\ref{sec:our_harmonization}) and standardization (\S\ref{sec:standard_latin}). During harmonization, we enforce consistency of arbitrary values to have fair training and evaluation. Standardization is more involved, where we change the grammatical system to be more Latin-specific. Both of these steps are done automatically and simultaneously through conversion scripts.

\subsection{Annotation Agreement Between UD and LASLA} \label{sec:annot_agreement_before}
\begin{table}[htbp]
    \centering
    \footnotesize
    \setlength{\tabcolsep}{2pt}
    \begin{tabular}{|l|l|r|}
        \hline
        \textbf{Author} & \textbf{Work} & \textbf{\# Dups} \\
        \hline
        Caesar & Gallic War & 1127 \\
        Cicero & De Officiis & 447 \\
        Cicero & In Catilinam & 118 \\
        Ovid & Metamorphoses & 0\setfootnotemark\label{dup_sents_ovid} \\
        Petronius & Satyricon & 407 \\
        Propertius & Elegies & 183 \\
        Sallust & Bellum Catilinae & 228 \\
        Tacitus & Historiae & 50 \\
        Vergil & Aeneid & 47 \\
        \hline
    \end{tabular}
    \caption{For the nine texts shared between LASLA and UD (collectively; specifically, Perseus and PROIEL), number of duplicate sentences.}
    \label{tab:duplicate_sentences}
\end{table}

\footnotetext[\getrefnumber{dup_sents_ovid}]{Ovid's \textit{Metamorphoses} appears in both treebanks, but they cover different books of the text.}

\noindent
UD and LASLA happen to have re-annotated many of the same sentences, which gives a way to analyze annotation agreement between the projects.
We detect sentences that appear in both datasets (\S\ref{sec:dup_sents}),
finding 2607 such duplicates across eight Classical texts (Table \ref{tab:duplicate_sentences}),
which may be an underestimate since our duplicate detector will miss cases where sentence segmentation or tokenization differ.

We calculate annotation agreement before and after harmonization and standardization on our reduced set of features (Table \ref{tab:same_toks_lasla_ud_harmonized}). Some features, such as \verb|Degree|, \verb|Tense|, and \verb|VerbForm|, have low agreement due to mismatches between their possible value sets in UD and in LASLA. Other features, such as \verb|Gender|, \verb|Person|, and UPOS have low agreement due to remaining annotation differences.\footnote{See appendix for agreement rates across all features (Table \ref{tab:same_toks_lasla_ud}) and a comprehensive overview of the feature inventories (Table \ref{tab:feature_comparison_ud_lasla}).
}


\subsection{Our Harmonization Efforts} \label{sec:our_harmonization}

\citet{gamba-zeman-2023-latin} have already performed the bulk of the harmonization necessary for the UD treebanks. However, we are additionally attempting to harmonize LASLA with the UD treebanks. 

\paragraph{Remaining inconsistencies we've harmonized.} We have found some remaining inconsistencies, both within the UD treebanks and between UD and LASLA. Usually, neither is incorrect in their annotation, but without normalization this will cause unfair evaluation. Thus, we enforce consistent, arbitrary values in these cases. See \S\ref{sec:arbitrary_value_list} for specifics.

\paragraph{Collapsing feature values.} Another issue we encountered is that some UD treebanks lack certain feature values that are present in the others. \citet{gamba-zeman-2023-latin} were aware of this issue, and chose not to harmonize these values in order to preserve as much information as possible. This is understandable, as these features may be of interest to researchers. However, for our purposes, we have collapsed certain feature values together in order to have fairer evaluation of models trained on different treebanks.

For UPOS (universal part of speech), we have collapsed INTJ into PART across all treebanks, since two UD treebanks (ITTB and LLCT) do not use INTJ, instead using the value PART.
Additionally, for \verb|Degree|, we have collapsed \verb|Degree=Pos| into \verb|Degree=None|, since LASLA is the only treebank to use \verb|Pos|. The distinction between \verb|Degree=Pos| and \verb|Degree=None| is debated.\footnote{See the UD documentation for Degree in Latin \hyperlink{https://universaldependencies.org/la/feat/Degree.html}{here}.}
We note that \citet{gamba-zeman-2023-latin} 
also collapsed \verb|Degree=Pos| and \verb|Degree=Dim| into \verb|Degree=None|, so this decision has precedent.

\subsection{Conversion to Standard Latin Grammar}\label{sec:standard_latin}
\begin{table}[htbp]
    \centering
    \footnotesize
    \setlength{\tabcolsep}{2pt}
    \begin{tabular}{l|rrr|rrr}
    \toprule
 & \multicolumn{3}{c|}{Before}  & \multicolumn{3}{c}{After} \\ 
Feature &  \% same & \# same & \# total &  \% same & \# same & \# total\\ 
\hline
Case & 97.8 & 20372 & 20821 & 97.8 & 20372 & 20821\\ 
Degree & 8.5 & 598 & 6998 & \textbf{69.5} & 598 & 860\\ 
Gender & 74.7 & 14965 & 20034 & \textbf{75.2} & 14964 & 19911\\ 
\ \ \emph{(loose)}
& 97.2 & 19481 & 20034 & \textbf{97.8} & 19477 & 19911\\ 
Mood & 99.4 & 5279 & 5312 & \textbf{97.3} & 8621 & 8864\\ 
VerbForm & 93.2 & 8264 & 8867 & -- & -- & --\\ 
Number & 97.9 & 25672 & 26211 & 97.9 & 25543 & 26088\\ 
Person & 91.0 & 6089 & 6692 & 91.0 & 6089 & 6692\\ 
Tense & 77.3 & 5228 & 6766 & \textbf{96.7} & 8184 & 8465\\ 
Voice & 96.0 & 7493 & 7809 & \textbf{96.5} & 8554 & 8864\\ 
UPOS & 93.0 & 34814 & 37425 & 93.0 & 34821 & 37425 \\
\bottomrule
    \end{tabular}
    \caption{Percent and number of tokens in the duplicate LASLA and Harmonized UD sentences that have the 
    exact same value  
    for each feature, before and after our harmonization and standardization. Percent is out of tokens that had a non-None value in either UD or LASLA. After our changes, Mood and VerbForm are collapsed into Mood only, but we list them separately before.
    Percentages after our changes are \textbf{boldfaced} when there is improved agreement. 
    }
    \label{tab:same_toks_lasla_ud_harmonized}
\end{table}
UD was developed with cross-linguistic goals in mind, offering a set of universal tags applicable to all languages. However, prior to the harmonization efforts by \citet{gamba-zeman-2023-latin}, many Latin UD treebanks employed standard Latin values for certain features, reflecting a long-standing desire for a more Latin-specific tagset. Harmonization and conversion to UD has relegated these Latin-specific values to a secondary status. This poses a key challenge for evaluation, as these two annotation styles are not comparable.
 
Although UD provides a valuable cross-linguistic framework, we believe Latin is also useful to study on its own, within long-standing approaches to Latin linguistics (e.g.\ \citealt{latin-grammar-textbook}). 
 The UD treebanks remain the most complete, high-quality source of morphological annotations for Latin. To bridge the gap between UD and standard Latin linguistics, we offer an alternative version that uses more standard Latin grammar.
 In particular, we standardize the treebanks to follow Pre-UD Perseus's (Table \ref{tab:datasets} row 1) features: UPOS, person, number, tense, mood, voice, gender, case, and degree. This set is nearly identical to \citet{burns2023latincy}'s, except that LatinCy separately predicts Mood and VerbForm (which we combine). For most of these features, UD has a corresponding feature that we can easily extract. The exceptions are Tense and Mood, where we developed a more elaborate method of standardization (\S\ref{sec:tense_mood}). For example, Latin tense traditionally has six possible values (present, imperfect, perfect, future, pluperfect, future perfect) which are standardized across pedagogical materials \cite{latin-grammar-textbook, wheelock2010wheelock}. However, UD's \verb|Tense| feature only includes four of these values (present, past, future, pluperfect), which is why we must perform a conversion. 

 We choose to convert to Standard Latin Grammar \textit{before} training, rather than perform postprocessing on the predictions of a model trained on the UD tagset, for two reasons: 1) preprocessing allows for more precise conversions based on known treebank sources, addressing inconsistencies between treebanks, and 2) model predictions may combine features from various annotation schemes and be grammatically inconsistent, making postprocessing complex and potentially unreliable. 

 \begin{figure}[t]
    \centering
    \includegraphics[width=\linewidth]{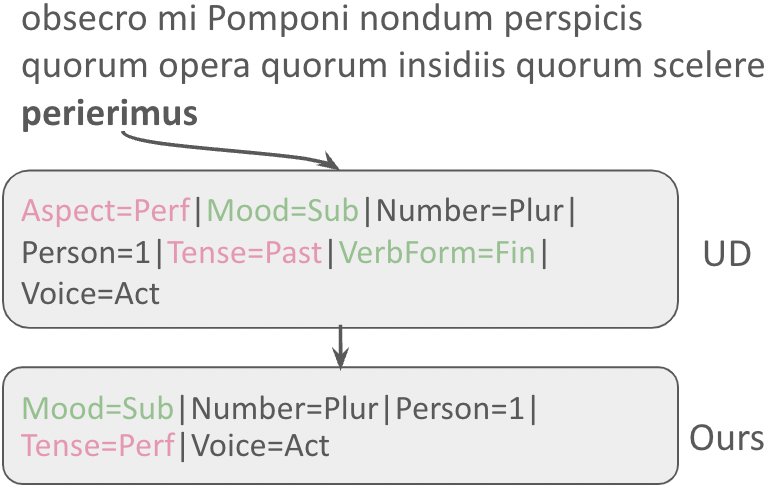}
    \caption{Example of how a token's set of morphological features changes after standardization, from Cicero’s \textit{Letters to Atticus} Book 3 Letter 9.}
    \label{fig:standard_grammar}
\end{figure}


\subsection{Remaining Inconsistencies} \label{sec:rem_inconsistencies}
After our harmonization and standardization, most features have high annotation agreement between LASLA and UD (Table \ref{tab:same_toks_lasla_ud_harmonized}). Degree and UPOS, two features that already had low agreement within UD (Table \ref{tab:same_toks_pers_proiel_harmonized}), saw improved but not high agreement after UD-LASLA harmonization. These are likely due to fundamental differences in the annotation process which may require reannotation to fix.

We modify how our models are trained to account for the two following differences (\S\ref{sec:bert_tagger}):
\begin{itemize}
    \item In LASLA, the \verb|Gender| feature can take multiple values to represent possible genders based only on the word form (disregarding the context of the sentence). In the UD treebanks, \verb|Gender| is assigned one value that depends on the sentence. This causes the low annotation agreement for \verb|Gender| in Table \ref{tab:same_toks_lasla_ud_harmonized}. If we use a looser criterion---counting the annotations as the same when the UD gender value matches one of LASLA's gender values---we do see higher agreement (\textit{(loose)} in Table \ref{tab:same_toks_lasla_ud_harmonized}).
    \item In LASLA, personal pronouns are annotated with \verb|Person=None|, but in the UD treebanks personal pronouns have non-\verb|None| values.\footnote{This will be simple to fix in future work, since there are limited personal pronoun lemmas in Latin.}
\end{itemize}
We list additional differences in \S\ref{sec:app_remaining_diffs}.


\subsection{Our Custom Data Splits} \label{sec:custom_splits}
\begin{table}[h]
    \centering
    \footnotesize
    \begin{tabular}{lrrr}
        \toprule
        Time & Train & Dev & Test \\
        \midrule
        Classical (UD) & 6524 & 201 & 1041   \\
        Classical (UD+LASLA) & 102498 & 201 & 1041 \\
        Bible & 10451 & 322 & 1021  \\
        Postclass & 32661 & 1010 & 5003  \\
        \bottomrule
    \end{tabular}
    \caption{Number of sentences in our proposed train, dev, and test splits}
    \label{tab:new_splits_overview}
\end{table}
We create new data splits to emulate EvaLatin's cross-time sub-task which evaluates models on texts of a different time period than what they are trained on. 
When creating train/test splits for each time period, we keep the following constraints in mind:
    \textbf{1)} Individual works should be within a single split. For example, Ovid's \textit{Metamorphoses} should only appear in either the train or test set, rather than having a random sample of sentences in the train set with the rest in the test set.
    \textbf{2)} Make sure the test set is large enough for reasonable statistical power. We specifically choose to have a minimum of 1000 sentences in each test set.
    \textbf{3)} Only evaluate on UD data and not LASLA. Due to some annotation differences (see \S\ref{sec:rem_inconsistencies}), UD treebanks have more complete information than LASLA. This is in contrast to EvaLatin campaigns which evaluate on subsets of LASLA.

To make our dev sets, we randomly sample 3\% of sentences from each work in the train sets, making sure that we never sample from LASLA or any UD sentences that also appear in LASLA.

Due to these constraints, we are unable to keep the original UD test sets. Since we want test sets for each time period, we must construct Classical-specific splits. Perseus, despite being largely Classical, is too small for effective training. PROIEL contains some Classical texts but is mostly comprised of Biblical texts. We separate the Biblical content and combine the Classical texts from both treebanks to ensure a sufficiently large Classical train set.
Due to the first constraint, we cannot use ITTB's original train/test splits since \textit{Summa Contra Gentiles} appears in both the train and test set.
To help meet our second constraint, we do not use UDante's and LLCT's original splits.

To achieve our third constraint, our Classical train set must include all works that appear in both LASLA and UD, shown in Table \ref{tab:duplicate_sentences}.
We want to test two scenarios: training with and without LASLA data. In order to have enough training sentences in the UD-only scenario, we treat the letters of Cicero's \textit{Letters to Atticus} as separate texts (i.e. that can be distributed across the Classical train and test set), even though this conflicts with our first constraint.

We include a detailed description of which works appear in our custom train and test sets in the Appendix (Table \ref{tab:custom_splits_by_work}).

\section{Related Work: Morphological Tagging}

There is a long history of work analyzing POS and morphological tagging of Latin \cite{eger-etal-2015-lexicon, eger-etal-2016-lemmatization, straka-strakova-2020-udpipe}. Our work follows recent trends of using transformer-based contextual representations.

Several recent papers have explored morphological tagging for Latin. As part of the 2022 EvaLatin feature identification task \cite{sprugnoli-etal-2022-overview-evalatin}, participants trained and tested on a subset of data from the LASLA corpus that had been automatically converted to UD format \cite{wrobel-nowak-2022-transformer-evalatin, mercelis-keersmaekers-2022-electra-evalatin}. Only a subset of UD morphological features were retained, partly to limit the task to morphological features identifiable by the word form
, and partly to avoid features affected by annotation differences. Participants were then able to train on combined UD and LASLA data if they wished, but models were only evaluated on EvaLatin test sets, not UD test sets.

\citet{nehrdich-hellwig-2022-accurate-dep-parsing} used LatinBERT \cite{bamman2020latin} to train a morphological tagger predicting the case, gender, number, tense and verbform features. Its outputs were then fed into the authors' dependency parser, outperforming prior work using UDPipe and static word embeddings \cite{straka2019evaluating}. Their training and test data came from three UD treebanks (ITTB, PROIEL, and Perseus). 

\citet{burns2023latincy} developed LatinCy, a full NLP pipeline for Latin which includes morphological feature classification.\footnote{Using SpaCy \cite{spacy2}} Notably, this pipeline was trained on all five UD treebanks with early attempts made at harmonization, using a smaller tagset than UD that is closer to standard analyses of Latin grammar (Table \ref{tab:datasets} row 3).
Recently, \citet{gamba-zeman-2023-latin} performed more rigorous harmonization of morphological features across the five UD Latin treebanks (Table \ref{tab:datasets} row 4). They reported accuracy before and after harmonization, training and testing on each pair of treebanks using fasttext embeddings \cite{grave-etal-2018-learning-fasttext} with UDPipe \cite{straka-etal-2016-udpipe} or Stanza \cite{qi-etal-2020-stanza}. Harmonization was shown to improve accuracy when training and testing on two different treebanks.

Part-of-speech (POS) tagging is closely related to morphological tagging. In the 2020 EvaLatin campaign, participants trained and tested POS taggers on a subset of the LASLA corpus
\cite{sprugnoli-etal-2020-overview-evalatin}. 
More recently, \citeauthor{riemenschneider-frank-2023-exploring-llm} pretrained a trilingual RoBERTa \cite{liu2019roberta} model on English, Ancient Greek, and Latin which surpassed the 2022 EvaLatin competitors 
(Table \ref{tab:datasets} row 6). 
Thus, the current SOTA models for Latin POS tagging are all transformer-based. Additionally, \citeauthor{riemenschneider-frank-2023-exploring-llm}'s trilingual model underperformed their monolingual Ancient Greek model, suggesting a monolingual Latin model could prove even stronger, given sufficient pretraining data.

Researchers have also experimented with using GPT3.5-Turbo and GPT4 for POS tagging of 16th century Latin texts \cite{stussi-strobel-2024-part-lat-gpt}. No POS-annotated data exists for 16th century Latin, so the authors experimented with zero-shot prompting and finetuning using data from the five UD treebanks. 
Although the UD testsets are not entirely comparable with EvaLatin's, the accuracy of these GPT-based approaches seems low when compared to the results of EvaLatin's POS tagging shared task.

Although substantial progress has been made in Latin morphological tagging, gaps still exist.
Aside from \citet{gamba-zeman-2023-latin} and \citet{burns2023latincy}, prior work has not leveraged all five UD treebanks for training \textit{and} evaluation.
While \citet{gamba-zeman-2023-latin} measure overall tagging accuracy,
more detailed analysis of specific morphological features and diachronic trends has been left to future work.
Moreover, to our knowledge no recent paper has evaluated the currently available taggers on UD test data.
\section{Experiments} \label{sec:experiments}

We use three metrics in our evaluations: whole string morphological accuracy, macro F1 for individual features, and F1 for particular feature-values. See \ref{sec:metrics} for more detailed explanations.

\subsection{Our LatinBERT-based Tagger}\label{sec:bert_tagger}
Following other recent working finding SOTA performance with transformer-based taggers \cite{sprugnoli-etal-2022-overview-evalatin, riemenschneider-frank-2023-exploring-llm}, we finetune a tagger on top of LatinBERT \cite{bamman2020latin}. Similar to \citet{riemenschneider-frank-2023-exploring-llm}, our tagger uses a separate classification head for every morphological feature, all trained simultaneously---a simple choice which could be improved upon in future work. 

When training on LASLA, we sometimes do not train a particular feature head based on a token's feature values. First, if \verb|Gender| has multiple values we do not train the Gender prediction head.  We want to keep our set of possible \verb|Gender| values limited to the standard three (\verb|Masc|, \verb|Fem|, and \verb|Neut|). 
Second, if the token is a personal pronoun and \verb|Person=None|, we do not train the Person prediction head. Having a null value here is inconsistent with the rest of our data. Since we do not know the true value, we choose not to train in this instance. 
If either of these two cases apply to a particular token, then that token will not contribute to the loss for either the Gender or Person classifier head. Other heads are unaffected.

\subsection{Comparison to Previous Taggers}
\begin{table}[]
    \centering
    \footnotesize
    \setlength{\tabcolsep}{2pt}
    \begin{tabularx}{\linewidth}{ll|*{5}{>{\raggedright\arraybackslash}X}}
    \toprule
        Model & Train & per- & pro- & llct & ittb & uda- \\
        & Set(s) & seus & iel & & & nte\\
        \hline
        LatinCy & All UD & .726 & .740 & .792 & .809 & .736 \\
        BERT & All UD & \textbf{.929} & .962 & .969 & .982 & \textbf{.910} \\ \hline
        Stanza & In-Domain UD & .787 & .929 & .969 & .965 & .819 \\
        BERT & In-Domain UD & .915 & \textbf{.962} & \textbf{.977} & \textbf{.984} & .903 \\
        \bottomrule
    \end{tabularx}
    \caption{Whole string accuracy of \textbf{morphological features}. Train set is either All 5 UD treebanks, or a single In-Domain UD Treebank (i.e., same as the Test column).
    }
    \label{tab:perf_original_morph_acc}
\end{table}
\begin{table}[]
    \centering
    \setlength{\tabcolsep}{2pt}
    \footnotesize
    \begin{tabularx}{\linewidth}{ll|*{5}{>{\raggedright\arraybackslash}X}}
    \toprule
Model & Metric & per- & pro- & llct & ittb & uda- \\
& & seus & iel & &  & nte \\ \hline
Stanza & POS Macro F1 & .072 & .253 & .284 & .227 & .122 \\
BERT & POS Macro F1 & \textbf{.066} & \textbf{.191} & \textbf{.185} & \textbf{.144} & \textbf{.101} \\ \hline
Stanza & Morph Acc & .058 & .179 & .275 & .177 & .077 \\
BERT & Morph Acc & \textbf{.016} & \textbf{.069} & \textbf{.186} & \textbf{.074} & \textbf{.030} \\
\bottomrule
    \end{tabularx}
    \caption{Average difference between in and out of domain performance, for each of the 5 UD treebank test sets (columns); this work (BERT rows) always attains a smaller difference.}
    \label{tab:cross_dom_perf}
\end{table}
In this section, we use the official UD train/test splits for comparison to previous work but converted to our harmonized and standardized tagset.
We compare our BERT taggers to two sets of taggers previously evaluated on UD data: LatinCy (Table \ref{tab:datasets} row 3) and five Stanza models trained on the five Harmonized UD treebanks (Table \ref{tab:datasets} row 4). LatinCy uses a non-transformer neural architecture as part of the SpaCy pipeline, along with static floret vectors \cite{Boyd_Warmerdam_2022}. Stanza has a Bi-LSTM architecture for its POS and morphological taggers and uses either word2vec \cite{zeman-etal-2018-conll} or fasttext \cite{10.1162/tacl_a_00051} embeddings, depending on the language.
For a fair comparison, we must convert between the different tagsets used by each tagger.
For LatinCy, rather than retraining the SpaCy pipeline ourselves, we convert its predictions on each official UD test set to our tagset.
This required little modification as LatinCy predicts a near-identical set of features and values.\footnote{LatinCy lacks two possible \texttt{Tense} values, \texttt{Perf} and \texttt{FutP}, which our tagset includes. In a more generous evaluation, where \texttt{Fut} and \texttt{Imp} are considered correct predictions for gold \texttt{FutP} and \texttt{Perf}, respectively, all morphological accuracy scores in Table \ref{tab:perf_original_morph_acc} increase by $\leq$ 5\%, with maximum accuracy on the LLCT test set at 0.826.} 
For the Stanza models, we retrain them on our harmonized and standardized versions of each UD training set (Table \ref{tab:datasets} row 8), since \citet{gamba-zeman-2023-universalising}'s models and their predictions are unreleased.
Replicating \citet{gamba-zeman-2023-universalising}, we only train the Stanza models on each individual treebank, rather than all UD data. We also use the same Latin fasttext embeddings \cite{grave-etal-2018-learning-fasttext} and default training parameters.

We report whole string morphological accuracy for each UD test set in Table \ref{tab:perf_original_morph_acc}.
Our BERT taggers consistently have the highest accuracy. The smallest treebanks, Perseus and UDante, see the most benefit from the BERT architecture and the out-of-domain training data.\footnote{We see similar trends for UPOS; see Table \ref{tab:perf_original_pos_f1}.}

%


%

When comparing two models' performance, we calculate statistical significance via randomized permutation testing (\citealt{Wasserman2004AllStats}).\footnote{As detailed in \S\ref{sec:p_testing}, we simply report $p$-values based on 10,000 null simulations; thus $p$=0 is possible and could be more conservatively interpreted as 
$p<.0003$ 
(``rule of three'': \citealt{eypasch1995rule3}).}
When comparing our All-UD model to LatinCy and our in-domain models to Stanza, all comparisons were significant ($p$=0) for both UPOS Macro-F1 and morphological accuracy, except for LLCT UPOS Macro-F1 ($p$=0.13). 
So, in nearly all cases our BERT taggers performed significantly better at both UPOS and morphological tagging than previously released taggers, when trained on all or in-domain data.

We also find that our BERT taggers are more robust to out-of-domain data than the Stanza taggers. In Table \ref{tab:cross_dom_perf}, for each UD test set, we report the average difference between the in-domain test performance (training and testing on the same treebank) and out-of-domain test performance (training on a different treebank). This difference is always lower for our BERT models than for the Stanza models, suggesting that BERT has better cross-domain performance than Stanza.

\subsection{Performance on Our Custom Splits}
\begin{table}[]
    \centering
    {\footnotesize
    \begin{tabularx}{2.6in}{l|ccc}
    \toprule
     Model & \multicolumn{1}{c}{classical} & \multicolumn{1}{c}{bible} & \multicolumn{1}{c}{postclass}\\
     \hline
     &  \multicolumn{3}{c}{UPOS Macro-F1} \\
    classical-ud & \textbf{0.964} & 0.937 & 0.864\\
    classical-all & 0.949 & 0.799 & 0.839\\
    bible & 0.868 & \textbf{0.976} & 0.834\\
    postclass & 0.866 & 0.920 & \textbf{0.980}\\
    all-ud-custom & 0.961 & 0.975 & 0.976\\
    all-both-custom & 0.948 & 0.964 & 0.980\\
    \hline 
     &  \multicolumn{3}{c}{Morph Accuracy} \\
    classical-ud & 0.946 & 0.936 & 0.905\\
    classical-all & 0.945 & 0.941 & 0.908\\
    bible & 0.914 & 0.956 & 0.885\\
    postclass & 0.916 & 0.931 & 0.973\\
    all-ud-custom & \textbf{0.946} & 0.956 & 0.973\\
    all-both-custom & 0.939 & \textbf{0.960} & \textbf{0.974} \\
    \bottomrule
    \end{tabularx}
    }
    \caption{Performance of our BERT-based taggers when evaluated on custom time-period test sets.
    \vspace{-1mm}}
    \label{tab:perf_custom}
\end{table}
In total, we train six models including four trained on the sets described in Table \ref{tab:new_splits_overview}. The other two models are \verb|all-ud-custom|, trained on the Classical (UD Only), Bible, and Postclass train sets; and \verb|all-both-custom|, trained on the Classical (UD+LASLA), Bible, and Postclass train sets.
Since our LatinBERT taggers outperform the other taggers, we limit our focus to these BERT-based taggers.
We find that it is generally unnecessary to train a period-specific model. As Table \ref{tab:perf_custom} shows, models trained on all time periods have only slightly reduced UPOS Macro F1 and have slightly increased morphological accuracy compared to the models trained on a single domain.


\paragraph{Addition of LASLA data boosts performance for some rare feature values, but decreases it for other features.}
Although there is only a slight difference in \textit{overall} morphological accuracy with the addition of LASLA data, F1 of particular feature-values improves. When evaluating the \verb|classical-ud| and \verb|classical-all| models on the Classical test set, 
F1 increases from 
0.907 to 0.941 for \verb|Case=Dat| ($p$=0.0028), 
and 0.800 to 0.909 for \verb|Mood=Ger| ($p$=0.0).
We also found that some features' Macro F1 scores decreased with the inclusion of LASLA.
This behavior is most prominent for Degree (0.96 to 0.91, $p$=0.0001) and UPOS (0.96 to 0.95, $p$=0.0).
Since the duplicate sentences in LASLA and UD have low annotation agreement for Degree and UPOS (Table \ref{tab:same_toks_lasla_ud_harmonized}), the addition of LASLA data likely led to noisier training labels for these two features.

\paragraph{Most errors involve acontextual ambiguity.}
We randomly sample 100 tokens whose morphology was predicted incorrectly by our \verb|all-ud-custom| model,\footnote{33 tokens from Classical texts, 33 from the bible, 12 from Aquinas' works, 11 from LLCT, 11 from Dante's works.} and annotated them according to six error types: illegal, lexical, genuine acontextual ambiguity, annotation differences, gold wrong, other.

\textbf{Illegal.} We found four illegal errors in which the model combined morphology and/or UPOS in a way that breaks rules of grammar.
Three of these involved the token \textit{quod}. For example, when the gold annotation labeled \textit{quod} as SCONJ, the model correctly predicted SCONJ but incorrectly predicted \verb|Gender=Neut| and \verb|Number=Sing|, when a SCONJ should have no value for those features. In the fourth case, when the gold was PRON, the model again correctly predicted PRON but incorrectly predicted \verb|Case=None| and \verb|Number=None|, even though a PRON should have values for those features.

\textbf{Lexical.} We found eight \textit{lexical errors} where the predicted combination of UPOS and morphological features is legal in general, but is impossible given the particular token based on lexical information. 
For example, let's consider the token \textit{ista} whose gold annotation is a DET with \verb|Case=Nom|, \verb|Gender=Fem|, and \verb|Num=Sing|. This word is a demonstrative adjective with 1st and 2nd declension endings, so out of context there are only a few combinations of morphological features possible: either 
\texttt{Case=Nom,Abl|Gender=Fem|Num=Sing}
or 
\texttt{Case=Nom,Acc|Gender=Neut|Num=Plur}.
Our tagger incorrectly predicted \verb|Case=Acc| but correctly predicted \verb|Gender=Fem| and \verb|Num=Sing|. Even though its predictions for Gender and Number are correct, they do not form a valid combination of feature values for this token.

\textbf{Genuine acontextual ambiguity.} Most errors (67) were due to \textit{genuine acontextual ambiguity}. This means that, out of context, the model's prediction is legal and valid given the particular token's lexical information, but in context it is incorrect. We would hope that BERT, as a contextual model, could still predict these cases correctly but it seems to struggle. 
\begin{figure}[t]
    \centering
    \includegraphics[width=\linewidth]{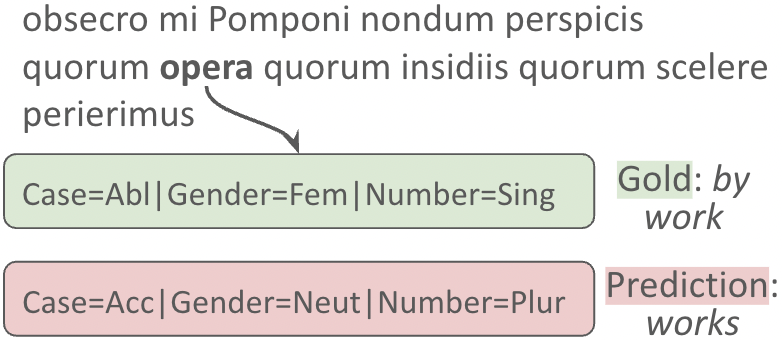}
    \caption{Example of an error in the model's prediction due to acontextual ambiguity, from Cicero’s \textit{Letters to Atticus} Book 3 Letter 9.
    \vspace{-1mm}}
    \label{fig:acontextual_error}
\end{figure}

Figure \ref{fig:acontextual_error} shows an example of this error type. In other contexts, the token \textit{opera} can be accusative plural, as the model predicted, but within this sentence it must be ablative singular. The verb \textit{perierimus} (we have been ruined) does not take an object, so \textit{opera} cannot be accusative. Additionally, the structure of \textit{quorum opera} is repeated with \textit{quorum insidiis} and \textit{quorum scelere}. The nouns \textit{insidiis} and \textit{scelere} are clearly ablative, suggesting that \textit{opera} should be the same case. This makes more sense contextually: \textit{perierimus} (we have been ruined) \textit{quorum opera} (by whose work).

\textbf{Annotation differences.} Nine errors were due to remaining annotation differences, discussed more thoroughly in \S\ref{sec:app_remaining_diffs}.


\textbf{Gold wrong.} Nine errors were caused by incorrect gold annotations. These include missing \verb|Case| value for nouns, and incorrect UPOS and morphological features.

\paragraph{Words segmented by the tokenizer have a higher error rate.}
Because of the presence of lexical errors in our model's predictions, we investigated whether the LatinBERT tokenizer segments words in a morphologically-aware manner.
We find that the majority (81\%) of words in our three custom test sets correspond to a single subtoken for the tokenizer.
For these word tokens, our \verb|all-ud-custom| model achieves 98.3\% accuracy on UPOS and 97.2\% accuracy on all morphological features.
In the case that word tokens are split into \textit{multiple} subtokens, performance degrades; Accuracy drops slightly for UPOS to 97.5\% and more dramatically for morphological features to 93.6\%.
Since most words are not segmented and those that are have worse performance, we hypothesize that the model is not able to learn Latin's inflections, which could hypothetically aid in the tagging of rarer words. The relationship between token frequency, word segmentation, and downstream performance is a promising direction for analysis in future work.
This aligns with previous findings for English that transformer models with WordPiece tokenizers have lower generalization ability than those with morphologically-aware tokenization \cite{hofmann-etal-2021-superbizarre}.

\section{Conclusion and Future Work}\label{sec:conclusion}
In this work, we consider the diverse time periods represented in the Latin treebanks when training and evaluating morphological taggers. We hope the genre metadata we provide can be used for future cross-genre analysis of Latin, similar to the cross-time analysis we present in this paper.

We also believe further improvements can be made through 
the harmonization of remaining annotation differences (\S\ref{sec:rem_inconsistencies}) and more informed modeling choices. Specifically, we hypothesize that (1) conditioning morphological feature prediction on UPOS, or vice versa; (2) enforcing grammatical constraints through modeling, rather than only through training data; and (3) constructing a morphologically-aware tokenizer may all lead to improved performance.

\section*{Acknowledgments}

We thank Gregory Crane, David Smith, Geoff Bakewell, the UMass NLP group, and anonymous reviewers for feedback.
This
work was supported by NSF CAREER 1845576.
Any opinions, findings, and conclusions or recommendations expressed in this material are those of
the authors and do not necessarily reflect the views
of the National Science Foundation.

\bibliography{acl_latex}

\begin{thebibliography}{42}
\providecommand{\natexlab}[1]{#1}

\bibitem[{Bamman and Burns(2020)}]{bamman2020latin}
David Bamman and Patrick~J. Burns. 2020.
\newblock \href {https://api.semanticscholar.org/CorpusID:221819449} {{Latin BERT: A Contextual Language Model for Classical Philology}}.
\newblock \emph{ArXiv}, abs/2009.10053.

\bibitem[{Bamman and Crane(2011)}]{perseus}
David Bamman and Gregory Crane. 2011.
\newblock The {A}ncient {G}reek and {L}atin {D}ependency {T}reebanks.
\newblock In \emph{Language Technology for Cultural Heritage}, pages 79--98, Berlin, Heidelberg. Springer Berlin Heidelberg.

\bibitem[{Bojanowski et~al.(2017)Bojanowski, Grave, Joulin, and Mikolov}]{10.1162/tacl_a_00051}
Piotr Bojanowski, Edouard Grave, Armand Joulin, and Tomas Mikolov. 2017.
\newblock \href {https://doi.org/10.1162/tacl_a_00051} {{Enriching Word Vectors with Subword Information}}.
\newblock \emph{Transactions of the Association for Computational Linguistics}, 5:135--146.

\bibitem[{Boyd and Warmerdam(2022)}]{Boyd_Warmerdam_2022}
Adriane Boyd and Vincent~D. Warmerdam. 2022.
\newblock \href {https://explosion.ai/blog/floret-vectors} {floret: lightweight, robust word vectors}.

\bibitem[{Burns(2023)}]{burns2023latincy}
Patrick~J. Burns. 2023.
\newblock \href {https://arxiv.org/abs/2305.04365} {{LatinCy: Synthetic Trained Pipelines for Latin NLP}}.
\newblock \emph{Preprint}, arXiv:2305.04365.

\bibitem[{Cecchini et~al.(2020{\natexlab{a}})Cecchini, Sprugnoli, Moretti, and Passarotti}]{udante}
Flavio Cecchini, Rachele Sprugnoli, Giovanni Moretti, and Marco Passarotti. 2020{\natexlab{a}}.
\newblock {UD}ante: {F}irst {S}teps {T}owards the {U}niversal {D}ependencies {T}reebank of {D}ante's {L}atin works.

\bibitem[{Cecchini et~al.(2020{\natexlab{b}})Cecchini, Korkiakangas, and Passarotti}]{llct}
Flavio~Massimiliano Cecchini, Timo Korkiakangas, and Marco Passarotti. 2020{\natexlab{b}}.
\newblock \href {https://aclanthology.org/2020.lrec-1.117} {{A New {L}atin Treebank for {U}niversal {D}ependencies: Charters between {A}ncient {L}atin and {R}omance Languages}}.
\newblock In \emph{Proceedings of the Twelfth Language Resources and Evaluation Conference}, pages 933--942, Marseille, France. European Language Resources Association.

\bibitem[{de~Marneffe et~al.(2021)de~Marneffe, Manning, Nivre, and Zeman}]{de-marneffe-etal-2021-universal}
Marie-Catherine de~Marneffe, Christopher~D. Manning, Joakim Nivre, and Daniel Zeman. 2021.
\newblock \href {https://doi.org/10.1162/coli_a_00402} {{U}niversal {D}ependencies}.
\newblock \emph{Computational Linguistics}, 47(2):255--308.

\bibitem[{Denooz(2004)}]{denooz2004opera-lasla}
Joseph Denooz. 2004.
\newblock {Opera Latina: une base de donn{\'e}es sur internet}.
\newblock \emph{Euphrosyne}, 32:79--88.

\bibitem[{Eger et~al.(2016)Eger, Gleim, and Mehler}]{eger-etal-2016-lemmatization}
Steffen Eger, R{\"u}diger Gleim, and Alexander Mehler. 2016.
\newblock \href {https://aclanthology.org/L16-1239} {{Lemmatization and Morphological Tagging in {G}erman and {L}atin: A Comparison and a Survey of the State-of-the-art}}.
\newblock In \emph{Proceedings of the Tenth International Conference on Language Resources and Evaluation ({LREC}'16)}, pages 1507--1513, Portoro{\v{z}}, Slovenia. European Language Resources Association (ELRA).

\bibitem[{Eger et~al.(2015)Eger, vor~der Br{\"u}ck, and Mehler}]{eger-etal-2015-lexicon}
Steffen Eger, Tim vor~der Br{\"u}ck, and Alexander Mehler. 2015.
\newblock \href {https://doi.org/10.18653/v1/W15-3716} {Lexicon-assisted tagging and lemmatization in {L}atin: A comparison of six taggers and two lemmatization methods}.
\newblock In \emph{Proceedings of the 9th {SIGHUM} Workshop on Language Technology for Cultural Heritage, Social Sciences, and Humanities ({L}a{T}e{CH})}, pages 105--113, Beijing, China. Association for Computational Linguistics.

\bibitem[{Eypasch et~al.(1995)Eypasch, Lefering, Kum, and Troidl}]{eypasch1995rule3}
Ernst Eypasch, Rolf Lefering, CK~Kum, and Hans Troidl. 1995.
\newblock Probability of adverse events that have not yet occurred: a statistical reminder.
\newblock \emph{{BMJ}}, 311(7005):619--620.

\bibitem[{Gamba and Zeman(2023{\natexlab{a}})}]{gamba-zeman-2023-latin}
Federica Gamba and Daniel Zeman. 2023{\natexlab{a}}.
\newblock \href {https://aclanthology.org/2023.alp-1.7} {{Latin Morphology through the Centuries: Ensuring Consistency for Better Language Processing}}.
\newblock In \emph{Proceedings of the Ancient Language Processing Workshop}, pages 59--67, Varna, Bulgaria. INCOMA Ltd., Shoumen, Bulgaria.

\bibitem[{Gamba and Zeman(2023{\natexlab{b}})}]{gamba-zeman-2023-universalising}
Federica Gamba and Daniel Zeman. 2023{\natexlab{b}}.
\newblock \href {https://aclanthology.org/2023.udw-1.2} {Universalising {L}atin {U}niversal {D}ependencies: a harmonisation of {L}atin treebanks in {UD}}.
\newblock In \emph{Proceedings of the Sixth Workshop on Universal Dependencies (UDW, GURT/SyntaxFest 2023)}, pages 7--16, Washington, D.C. Association for Computational Linguistics.

\bibitem[{Grave et~al.(2018)Grave, Bojanowski, Gupta, Joulin, and Mikolov}]{grave-etal-2018-learning-fasttext}
Edouard Grave, Piotr Bojanowski, Prakhar Gupta, Armand Joulin, and Tomas Mikolov. 2018.
\newblock \href {https://aclanthology.org/L18-1550} {{Learning Word Vectors for 157 Languages}}.
\newblock In \emph{Proceedings of the Eleventh International Conference on Language Resources and Evaluation ({LREC} 2018)}, Miyazaki, Japan. European Language Resources Association (ELRA).

\bibitem[{Greene and Resnik(2009)}]{greene-resnik-2009-words}
Stephan Greene and Philip Resnik. 2009.
\newblock \href {https://aclanthology.org/N09-1057} {{More than Words: Syntactic Packaging and Implicit Sentiment}}.
\newblock In \emph{Proceedings of Human Language Technologies: The 2009 Annual Conference of the North {A}merican Chapter of the Association for Computational Linguistics}, pages 503--511, Boulder, Colorado. Association for Computational Linguistics.

\bibitem[{Greenough and Allen(1903)}]{latin-grammar-textbook}
James~B. Greenough and Joseph~Henry Allen. 1903.
\newblock \href {https://archive.org/details/allengreenoughsn00alleiala/mode/2up} {\emph{Allen and Greenough's New Latin Grammar for Schools and Colleges}}.
\newblock Ginn \& Company.

\bibitem[{Haug and J{\o}hndal(2008)}]{proiel}
Dag Trygve~Truslew Haug and Marius~L. J{\o}hndal. 2008.
\newblock \href {https://api.semanticscholar.org/CorpusID:204978005} {Creating a {P}arallel {T}reebank of the {O}ld {I}ndo-{E}uropean {B}ible{T}ranslations}.

\bibitem[{Hofmann et~al.(2021)Hofmann, Pierrehumbert, and Sch{\"u}tze}]{hofmann-etal-2021-superbizarre}
Valentin Hofmann, Janet Pierrehumbert, and Hinrich Sch{\"u}tze. 2021.
\newblock \href {https://doi.org/10.18653/v1/2021.acl-long.279} {{Superbizarre Is Not Superb: Derivational Morphology Improves {BERT}{'}s Interpretation of Complex Words}}.
\newblock In \emph{Proceedings of the 59th Annual Meeting of the Association for Computational Linguistics and the 11th International Joint Conference on Natural Language Processing (Volume 1: Long Papers)}, pages 3594--3608, Online. Association for Computational Linguistics.

\bibitem[{Honnibal and Montani(2017)}]{spacy2}
Matthew Honnibal and Ines Montani. 2017.
\newblock {spaCy 2}: Natural language understanding with {B}loom embeddings, convolutional neural networks and incremental parsing.
\newblock To appear.

\bibitem[{Johnson et~al.(2021)Johnson, Burns, Stewart, Cook, Besnier, and Mattingly}]{johnson-etal-2021-classical}
Kyle~P. Johnson, Patrick~J. Burns, John Stewart, Todd Cook, Cl{\'e}ment Besnier, and William J.~B. Mattingly. 2021.
\newblock \href {https://doi.org/10.18653/v1/2021.acl-demo.3} {{The {C}lassical {L}anguage {T}oolkit: {A}n {NLP} Framework for Pre-Modern Languages}}.
\newblock In \emph{Proceedings of the 59th Annual Meeting of the Association for Computational Linguistics and the 11th International Joint Conference on Natural Language Processing: System Demonstrations}, pages 20--29, Online. Association for Computational Linguistics.

\bibitem[{Korkiakangas and Passarotti(2011)}]{llct1}
Timo Korkiakangas and Marco~Carlo Passarotti. 2011.
\newblock \href {https://api.semanticscholar.org/CorpusID:17711435} {{Challenges in Annotating Medieval Latin Charters}}.
\newblock \emph{J. Lang. Technol. Comput. Linguistics}, 26:103--114.

\bibitem[{Liu et~al.(2019)Liu, Ott, Goyal, Du, Joshi, Chen, Levy, Lewis, Zettlemoyer, and Stoyanov}]{liu2019roberta}
Yinhan Liu, Myle Ott, Naman Goyal, Jingfei Du, Mandar Joshi, Danqi Chen, Omer Levy, Mike Lewis, Luke Zettlemoyer, and Veselin Stoyanov. 2019.
\newblock \href {https://arxiv.org/abs/1907.11692} {{RoBERTa: A Robustly Optimized BERT Pretraining Approach}}.
\newblock \emph{Preprint}, arXiv:1907.11692.

\bibitem[{Mercelis and Keersmaekers(2022)}]{mercelis-keersmaekers-2022-electra-evalatin}
Wouter Mercelis and Alek Keersmaekers. 2022.
\newblock \href {https://aclanthology.org/2022.lt4hala-1.30} {{An {ELECTRA} Model for {L}atin Token Tagging Tasks}}.
\newblock In \emph{Proceedings of the Second Workshop on Language Technologies for Historical and Ancient Languages}, pages 189--192, Marseille, France. European Language Resources Association.

\bibitem[{Nehrdich and Hellwig(2022)}]{nehrdich-hellwig-2022-accurate-dep-parsing}
Sebastian Nehrdich and Oliver Hellwig. 2022.
\newblock \href {https://aclanthology.org/2022.lt4hala-1.3} {{Accurate Dependency Parsing and Tagging of {L}atin}}.
\newblock In \emph{Proceedings of the Second Workshop on Language Technologies for Historical and Ancient Languages}, pages 20--25, Marseille, France. European Language Resources Association.

\bibitem[{Nunn(1922)}]{Nunn}
H.~P.~V. Nunn. 1922.
\newblock \href {https://archive.org/details/introductiontoec00nunnuoft/page/n15/mode/2up} {\emph{An introduction to ecclesiastical Latin}}, page x–xi.
\newblock Cambridge University Press.

\bibitem[{Passarotti(2019)}]{ittb}
Marco Passarotti. 2019.
\newblock \href {https://doi.org/doi:10.1515/9783110599572-017} {\emph{The Project of the Index Thomisticus Treebank}}, pages 299--320.
\newblock De Gruyter Saur, Berlin, Boston.

\bibitem[{Qi et~al.(2020)Qi, Zhang, Zhang, Bolton, and Manning}]{qi-etal-2020-stanza}
Peng Qi, Yuhao Zhang, Yuhui Zhang, Jason Bolton, and Christopher~D. Manning. 2020.
\newblock \href {https://doi.org/10.18653/v1/2020.acl-demos.14} {{{S}tanza: A Python Natural Language Processing Toolkit for Many Human Languages}}.
\newblock In \emph{Proceedings of the 58th Annual Meeting of the Association for Computational Linguistics: System Demonstrations}, pages 101--108, Online. Association for Computational Linguistics.

\bibitem[{Rashkin et~al.(2017)Rashkin, Bell, Choi, and Volkova}]{rashkin-etal-2017-multilingual}
Hannah Rashkin, Eric Bell, Yejin Choi, and Svitlana Volkova. 2017.
\newblock \href {https://doi.org/10.18653/v1/P17-2073} {{Multilingual Connotation Frames: A Case Study on Social Media for Targeted Sentiment Analysis and Forecast}}.
\newblock In \emph{Proceedings of the 55th Annual Meeting of the Association for Computational Linguistics (Volume 2: Short Papers)}, pages 459--464, Vancouver, Canada. Association for Computational Linguistics.

\bibitem[{Riemenschneider and Frank(2023)}]{riemenschneider-frank-2023-exploring-llm}
Frederick Riemenschneider and Anette Frank. 2023.
\newblock \href {https://doi.org/10.18653/v1/2023.acl-long.846} {{Exploring Large Language Models for Classical Philology}}.
\newblock In \emph{Proceedings of the 61st Annual Meeting of the Association for Computational Linguistics (Volume 1: Long Papers)}, pages 15181--15199, Toronto, Canada. Association for Computational Linguistics.

\bibitem[{Sala and Posner(2024)}]{brittanica-latin-language}
Marius Sala and Rebecca Posner. 2024.
\newblock \href {https://www.britannica.com/topic/Latin-language} {Latin language}.
\newblock In \emph{Encyclopedia Britannica}.

\bibitem[{Sap et~al.(2017)Sap, Prasettio, Holtzman, Rashkin, and Choi}]{sap-etal-2017-connotation}
Maarten Sap, Marcella~Cindy Prasettio, Ari Holtzman, Hannah Rashkin, and Yejin Choi. 2017.
\newblock \href {https://doi.org/10.18653/v1/D17-1247} {{Connotation Frames of Power and Agency in Modern Films}}.
\newblock In \emph{Proceedings of the 2017 Conference on Empirical Methods in Natural Language Processing}, pages 2329--2334, Copenhagen, Denmark. Association for Computational Linguistics.

\bibitem[{Sprugnoli et~al.(2022)Sprugnoli, Passarotti, Cecchini, Fantoli, and Moretti}]{sprugnoli-etal-2022-overview-evalatin}
Rachele Sprugnoli, Marco Passarotti, Flavio~Massimiliano Cecchini, Margherita Fantoli, and Giovanni Moretti. 2022.
\newblock \href {https://aclanthology.org/2022.lt4hala-1.29} {{Overview of the {E}va{L}atin 2022 Evaluation Campaign}}.
\newblock In \emph{Proceedings of the Second Workshop on Language Technologies for Historical and Ancient Languages}, pages 183--188, Marseille, France. European Language Resources Association.

\bibitem[{Sprugnoli et~al.(2020)Sprugnoli, Passarotti, Cecchini, and Pellegrini}]{sprugnoli-etal-2020-overview-evalatin}
Rachele Sprugnoli, Marco Passarotti, Flavio~Massimiliano Cecchini, and Matteo Pellegrini. 2020.
\newblock \href {https://aclanthology.org/2020.lt4hala-1.16} {{Overview of the {E}va{L}atin 2020 Evaluation Campaign}}.
\newblock In \emph{Proceedings of LT4HALA 2020 - 1st Workshop on Language Technologies for Historical and Ancient Languages}, pages 105--110, Marseille, France. European Language Resources Association (ELRA).

\bibitem[{Straka et~al.(2016)Straka, Haji{\v{c}}, and Strakov{\'a}}]{straka-etal-2016-udpipe}
Milan Straka, Jan Haji{\v{c}}, and Jana Strakov{\'a}. 2016.
\newblock \href {https://aclanthology.org/L16-1680} {{{UDP}ipe: Trainable Pipeline for Processing {C}o{NLL}-{U} Files Performing Tokenization, Morphological Analysis, {POS} Tagging and Parsing}}.
\newblock In \emph{Proceedings of the Tenth International Conference on Language Resources and Evaluation ({LREC}'16)}, pages 4290--4297, Portoro{\v{z}}, Slovenia. European Language Resources Association (ELRA).

\bibitem[{Straka and Strakov{\'a}(2020)}]{straka-strakova-2020-udpipe}
Milan Straka and Jana Strakov{\'a}. 2020.
\newblock \href {https://aclanthology.org/2020.lt4hala-1.20} {{{UDP}ipe at {E}va{L}atin 2020: Contextualized Embeddings and Treebank Embeddings}}.
\newblock In \emph{Proceedings of LT4HALA 2020 - 1st Workshop on Language Technologies for Historical and Ancient Languages}, pages 124--129, Marseille, France. European Language Resources Association (ELRA).

\bibitem[{Straka et~al.(2019)Straka, Straková, and Hajič}]{straka2019evaluating}
Milan Straka, Jana Straková, and Jan Hajič. 2019.
\newblock \href {https://arxiv.org/abs/1908.07448} {{Evaluating Contextualized Embeddings on 54 Languages in POS Tagging, Lemmatization and Dependency Parsing}}.
\newblock \emph{Preprint}, arXiv:1908.07448.

\bibitem[{St{\"u}ssi and Str{\"o}bel(2024)}]{stussi-strobel-2024-part-lat-gpt}
Elina St{\"u}ssi and Phillip Str{\"o}bel. 2024.
\newblock \href {https://aclanthology.org/2024.latechclfl-1.18} {{Part-of-Speech Tagging of 16th-Century {L}atin with {GPT}}}.
\newblock In \emph{Proceedings of the 8th Joint SIGHUM Workshop on Computational Linguistics for Cultural Heritage, Social Sciences, Humanities and Literature (LaTeCH-CLfL 2024)}, pages 196--206, St. Julians, Malta. Association for Computational Linguistics.

\bibitem[{Wasserman(2004)}]{Wasserman2004AllStats}
Larry Wasserman. 2004.
\newblock \emph{All of Statistics}.
\newblock Springer Science \& Business Media.

\bibitem[{Wheelock and LeFleur(2010)}]{wheelock2010wheelock}
F.M. Wheelock and R.A. LeFleur. 2010.
\newblock \href {https://books.google.com/books?id=ri8sqWAVnoMC} {\emph{Wheelock's Latin: The Classic Introductory Latin Course, Based on Ancient Authors}}.
\newblock Elsie Giddings Helman Memorial. HarperCollins.

\bibitem[{Wr{\'o}bel and Nowak(2022)}]{wrobel-nowak-2022-transformer-evalatin}
Krzysztof Wr{\'o}bel and Krzysztof Nowak. 2022.
\newblock \href {https://aclanthology.org/2022.lt4hala-1.31} {{Transformer-based Part-of-Speech Tagging and Lemmatization for {L}atin}}.
\newblock In \emph{Proceedings of the Second Workshop on Language Technologies for Historical and Ancient Languages}, pages 193--197, Marseille, France. European Language Resources Association.

\bibitem[{Zeman et~al.(2018)Zeman, Haji{\v{c}}, Popel, Potthast, Straka, Ginter, Nivre, and Petrov}]{zeman-etal-2018-conll}
Daniel Zeman, Jan Haji{\v{c}}, Martin Popel, Martin Potthast, Milan Straka, Filip Ginter, Joakim Nivre, and Slav Petrov. 2018.
\newblock \href {https://doi.org/10.18653/v1/K18-2001} {{{C}o{NLL} 2018 Shared Task: Multilingual Parsing from Raw Text to {U}niversal {D}ependencies}}.
\newblock In \emph{Proceedings of the {C}o{NLL} 2018 Shared Task: Multilingual Parsing from Raw Text to Universal Dependencies}, pages 1--21, Brussels, Belgium. Association for Computational Linguistics.

\end{thebibliography}

\appendix

\section{Appendix}
\label{sec:appendix}


\subsection{UD Genres}\label{sec:ud_genres_appendix}
We mark the following 12 genres: narrative, poem, short poem, letter, epic, history, satire, speech, treatise, Christian, Bible, legal.

These genres are not exclusive, so each text will have at least one, but possibly more genres marked.

In Figure \ref{fig:timeline_genre}, for simplicity we showed a subset of genres which are mutually exclusive. This also ensures the number of sentences shown in the figure exactly matches the number of sentences that exist in the UD treebanks. The additional genres that we left out of the figure are broader, covering multiple sub-genres. Specifically, \textit{narratives} includes some (not all) texts from every genre except for legal texts and speeches. \textit{Poems} includes epics and short poems, the two of which are mutually exclusive. \textit{Christian} includes the Bible itself, as well as the religious treatises of Thomas Aquinas.

\subsection{Finding Duplicate Sentences in LASLA and UD Treebanks} \label{sec:dup_sents} 
In order to detect \emph{duplicate sentences} between the treebanks, we first normalize the orthographic variation across the UD treebanks and LASLA. We used CLTK's \cite{johnson-etal-2021-classical} \href{https://docs.cltk.org/en/latest/_modules/cltk/alphabet/lat.html\#JVReplacer}{JV replacer}  on the harmonized UD treebanks, since LASLA's texts do not use the letters `j' or `v'. We also remove any punctuation present in the UD treebanks, as LASLA does not have punctuation. 

We search for duplicate sentences by finding sentence pairs with exact character or token overlap at the beginning or end of each sentence.

Within the duplicate sentences, we identify \emph{duplicate tokens} by searching for the longest overlapping, contiguous subsequence of tokens of each sentence. We search for exact token matches. Our reported number of duplicate tokens is an underestimate, since there are sometimes token mismatches within sentences that are genuine duplicates. For example, one sentence may have a numeral where the other has the word form of the number.

\subsection{Annotation Agreement }
\begin{table}[htbp]
    \centering
    \footnotesize
    \setlength{\tabcolsep}{3pt}
    \begin{tabular}{|l|r|r|r|}
        \hline
         \textbf{Feature} & \textbf{\% same} & \textbf{\# same} & \textbf{Total}\\
          \hline
            Case & 96.1 & 794 & 826 \\
            Degree & 50.0 & 5 & 10 \\
            Gender & 94.7 & 767 & 810 \\
            Mood & 93.1 & 349 & 375 \\
            Number & 97.7 & 1097 & 1123 \\
            Person & 100.0 & 347 & 347 \\
            Tense & 95.4 & 356 & 373 \\
            Voice & 98.4 & 369 & 375 \\
            UPOS & 97.6 & 1538 & 1576 \\
        \hline
    \end{tabular}
    \caption{Percent and number of tokens in the duplicate Perseus and PROIEL sentences that have the \textbf{exact same value} for each feature, after our harmonization and conversion to Standard Latin grammar}
    \label{tab:same_toks_pers_proiel_harmonized}
\end{table}
\begin{table}[htbp]
    \centering
    \footnotesize
    \setlength{\tabcolsep}{3pt}
    \begin{tabular}{|l|r|r|r|}
        \hline
         \textbf{Feature} & \textbf{\% same} & \textbf{\# same} & \textbf{Total}\\
          \hline
            AdpType & 76.8 & 2115 & 2753\\
            AdvType & 0.0 & 0 & 357\\
            Aspect & 97.1 & 8608 & 8864\\
            Case & 97.8 & 20372 & 20821\\
            Compound & 0.0 & 0 & 1\\
            ConjType & 0.0 & 0 & 5\\
            Degree & 8.5 & 598 & 6998\\
            Foreign & 0.0 & 0 & 2\\
            Gender & 74.7 & 14965 & 20034\\
            Gender\_loose & 97.2 & 19481 & 20034\\
            InflClass & 0.0 & 0 & 27580\\
            InflClass[nominal] & 0.0 & 0 & 3394\\
            Mood & 99.4 & 5279 & 5312\\
            Number & 97.9 & 25672 & 26211\\
            Number[psor] & 100.0 & 281 & 281\\
            NumForm & 0.0 & 0 & 268\\
            NumType & 71.2 & 497 & 698\\
            PartType & 6.2 & 4 & 65\\
            Person & 91.0 & 6089 & 6692\\
            Person[psor] & 96.3 & 501 & 520\\
            Polarity & 35.1 & 267 & 760\\
            Poss & 96.3 & 501 & 520\\
            PronType & 78.2 & 4952 & 6333\\
            Reflex & 91.7 & 584 & 637\\
            Tense & 77.3 & 5228 & 6766\\
            Variant & 0.0 & 0 & 43\\
            VerbForm & 93.2 & 8264 & 8867\\
            Voice & 96.0 & 7493 & 7809\\
            UPOS & 93.0 & 34814 & 37425\\
        \hline
    \end{tabular}
    \caption{Percent and number of tokens in the duplicate LASLA and Harmonized UD sentences that have the \textbf{exact same value} for each feature, before any harmonization or standardization by us.}
    \label{tab:same_toks_lasla_ud}
\end{table}
In Table \ref{tab:same_toks_pers_proiel_harmonized}, we show the annotation agreement between duplicate sentences in Perseus and PROIEL after our standardization and harmonization. Notably, these are both (Harmonized) UD treebanks (Table \ref{tab:datasets} row 4), and some annotation differences still remain, although agreement is generally still higher than between UD and LASLA.

In Table \ref{tab:same_toks_lasla_ud}, we show the annotation agreement between Harmonized UD and LASLA, before our harmonization and standardization. This table also shows the union of UD and LASLA's feature sets. There are many features we did not consider which could benefit from harmonization.

Finally, Table \ref{tab:feature_comparison_ud_lasla} is a venn diagram showing all possible features and values in UD and LASLA, before our harmonization and standardization. 

\subsection{Remaining Inconsistencies We've Harmonized} \label{sec:arbitrary_value_list}
Here is a full list of arbitrary values we've enforced for certain grammatical constructions. For each of these items, there are arguably multiple correct ways to annotate---and the Latin treebanks were annotating these differently.
\begin{itemize}
    \item Gerunds, Infinitives, and Supines should have \verb|Number=None|
    \item Gerunds, Gerundives, and Supines should have \verb|Tense=None|
    \item If UPOS is AUX, then \verb|Voice=Act|. This almost entirely applies to forms of \textit{sum}.
    \item Gerunds should have \verb|Voice=Act|, and Gerundives should have \verb|Voice=Pass|.
    \item Supines have \verb|Voice=Act|, unless used in a construction with \textit{iri}, in which case \verb|Voice=Pass|.
    \item Gerunds, Infinitives, and Supines should have \verb|Gender=None|.
\end{itemize}

\subsection{Standardizing Tense and Mood} \label{sec:tense_mood}
\textbf{Tense} We use the \texttt{TraditionalTense} field of the harmonized treebanks \cite{gamba-zeman-2023-latin}, rather than the UD approach to tense. Altogether, the UD Latin treebanks include four tenses (present, past, future, pluperfect) and four aspects (imperfective, perfective, prospective, inchoative). When tense and aspect are considered together, they can represent the seven traditional Latin tenses. However, this is less intuitive for Classicists or those whose goal is to study only Latin. We chose to revert back to the traditional tenses. We were able to use the \texttt{TraditionalTense} field for most tags, but to differentiate between future and future perfect it is also necessary to look at \texttt{Aspect}. Additionally, we found that infinitives did not have a \verb|TraditionalTense|, so we looked to the \verb|Aspect| feature value to determine the tense of infinitives.

 For LASLA, since it does not have a \verb|TraditionalTense| field, we look at both \verb|Tense| and \verb|Aspect| feature values to determine tense.

Our final set of tenses is: present, imperfect, perfect, pluperfect, future, and future perfect.

\textbf{Mood} Similar to tense, the non-finite moods are represented by a combination of the \texttt{Mood} and \texttt{VerbForm} fields in \citet{gamba-zeman-2023-latin}'s harmonized treebanks, with references to Latin-specific constructions being moved to the \texttt{TraditionalMood} field. Strictly speaking, non-finite verbs do not have mood, but traditional Latin grammars still classify the different non-finite verb-forms as "mood."\footnote{This is explained in the EvaLatin 2022 guidelines: \url{https://github.com/CIRCSE/LT4HALA/blob/master/2022/data_and_doc/EvaLatin_2022_guidelines_v1.pdf}}
Again, we opt to use the traditional terminology and follow the same tagset as the Perseus treebank. For finite verbs, this includes indicative, subjunctive, imperative; and for non-finite verbs, infinitive, participle, gerund, gerundive, and supine. 

For LASLA, we are able to take the mood directly from the \verb|Mood| feature for finite verbs, and from \verb|VerbForm| feature for non-finite verbs. This is because LASLA uses the Latin-specific \verb|Ger,Gdv,Sup| values for \verb|VerbForm|, unlike the harmonized UD treebanks.

\subsection{Remaining Inconsistencies We're Unable to Harmonize} \label{sec:app_remaining_diffs}
We are aware of the following differences, but leave their harmonization to future work:
\begin{itemize}
  \item The pre-UD Perseus treebank (Table \ref{tab:datasets} row 1) has an additional \verb|Voice| value for deponent verbs. After \citet{gamba-zeman-2023-latin}'s harmonization, deponent verbs in Perseus always have \verb|Voice=Act|, but deponent verbs in every other UD treebank have \verb|Voice=Pass|. We would like a system more similar to pre-UD Perseus with an additional \verb|Voice=Dep| value.
    \item ITTB is the only treebank that sometimes marks \textit{esse}, the infinitive of \textit{sum}, as NOUN with \verb|Mood=None|. 
\end{itemize}

The following annotation differences were found to cause 9\% of sampled errors in our BERT tagger's morphological predictions:
\begin{itemize}
    \item Whether to have \verb|Case=None| for undeclined nouns. 
    \item Whether deponent verbs should be labeled as \verb|Voice=Act| or \verb|Voice=Pass|. 
    \item Whether infinitives should have a value for \verb|Case|.
    \item Whether infinitives can have their UPOS be NOUN, \verb|Mood=None|, and \verb|Tense=None|. 
    \item Whether the pronoun \textit{sui} should always have \verb|Number=None|.
\end{itemize}

\subsection{Finetuning Details}
We use the same hyperparameters that \citet{bamman2020latin} used to finetune a POS tagger: Adam optimizer with learning rate $5 \times 10^{-5}$, early stopping patience of 10 epochs, batch size 32, dropout rate 0.25.
We keep the model with the lowest validation loss across all epochs. 

\subsection{Metrics}\label{sec:metrics}

\begin{table}[]
    \centering
    \footnotesize
    \setlength{\tabcolsep}{2pt}
    \begin{tabularx}{\linewidth}{ll|*{5}{>{\raggedright\arraybackslash}X}}
    \toprule
        Model & Train & per- & pro- & llct & ittb & uda- \\
        & Set(s) & seus & iel & & & nte\\
        \hline
LatinCy & All UD & .729 & .800 & .800 & .786 & .737 \\
BERT & All UD & \textbf{.872} & .974 & .982 & .980 & \textbf{.855} \\ \hline
Stanza & In-Domain UD & .809 & .967 & .982 & .977 & .841 \\
BERT & In-Domain UD & .867 & \textbf{.977} & \textbf{.984} & \textbf{.986} & .880 \\
        \bottomrule
    \end{tabularx}
    \caption{Macro F1 of \textbf{UPOS}. Train set is either All 5 UD treebanks, or a single In-Domain UD Treebank (i.e., same as the Test column).
    }
    \label{tab:perf_original_pos_f1}
\end{table}
\textbf{Whole-String Morphological Accuracy} Following the convention of \citet{gamba-zeman-2023-latin} and \citet{sprugnoli-etal-2022-overview-evalatin}, we consider the model’s prediction correct when every morphological feature is correctly predicted. We construct a morphological feature string from the predicted feature set, making sure to sort the features alphabetically. Then, we can test whether the predicted morphological string is an exact match to the gold string. Although this is a strict criteria, it indicates whether the model understands how all the morphological features fit together.

\textbf{Macro F1 for Individual Features} For UPOS and each individual morphological feature, we report Macro F1 in order to see how the model performs on rare feature values. If we define $F$ as a particular feature and $\mathcal{V}_F = \{v_1,...,v_n \}$ as the set of possible values that $F$ can take, then macro F1 is defined as $\frac{1}{n} \sum_{i=1}^n \text{F1}(v_i)$. Note that $v = $ \verb|None| is a possible value for every morphological feature, and is included in our calculation. 

\subsection{Randomized Permutation Testing} \label{sec:p_testing}
 Within a  null simulation, for each test set sentence we shuffle the two models' predictions, and store the absolute difference in the performance metric calculated from the entire shuffled test set.  We finally report the $p$-value as the fraction of 10,000 simulated absolute differences that are larger than the observed absolute difference. $p$=0 simply means the observed difference is larger than in all simulations; it could be more conservatively interpreted as $p<.0003$ \citep{eypasch1995rule3} due to Monte Carlo error.

\begin{table}[]
    \centering
    \footnotesize
    \begin{tabular}{l|rrr}
    \toprule
    Feature & classical & bible & postclass \\
    \hline
    Case & 0.946 & \textbf{0.953} & 0.948 \\
    Degree & 0.977 & \textbf{0.987} & 0.965 \\
    Gender & 0.968 & 0.977 & \textbf{0.982} \\
    Mood & 0.859 & 0.938 & \textbf{0.982} \\
    Number & 0.987 & 0.988 & 0.992 \\
    Person & 0.994 & 0.993 & 0.992 \\
    Tense & 0.955 & 0.977 & 0.954 \\
    Voice & 0.969 & 0.973 & 0.990 \\
    \bottomrule
    \end{tabular}
    \caption{Macro f1 of each individual feature for the \texttt{all-ud-custom} model. Note that macro f1 for \texttt{Mood} on the Classical test set seems low (0.859) because the model never predicts \texttt{Mood=Sup} (supine). Excluding that value, its macro f1 is 0.967.}
    \label{tab:all_morph_f1}
\end{table}

\begin{table*}[htbp]
    \centering
    \footnotesize
    \begin{tabular}{|p{3cm}|p{3.5cm}|p{3.5cm}|p{3.5cm}|}
        \toprule
        \textbf{Feature Union} & \textbf{UD Only Values} & \textbf{Value Intersection} & \textbf{LASLA Only Values} \\
        \hline
        Abbr &  & Yes &  \\\hline
        AdpType & Post & Prep &  \\\hline
        AdvType & Loc, Tim &  &  \\\hline
        Aspect & Inch & Imp, Perf, Prosp &  \\\hline
        Case &  & Loc, Acc, Abl, Voc, Nom, Dat, Gen &  \\\hline
        Compound & Yes &  &  \\\hline
        ConjType &  &  & Cmpr \\\hline
        Degree & Dim & Abs, Cmp & Pos \\\hline
        Foreign &  & Yes &  \\\hline
        Form & Emp &  &  \\\hline
        Gender &  & Fem, Neut, Masc & Fem,Neut, Fem,Masc,Neut, Fem,Masc, Masc,Neut \\\hline
        InflClass &  & LatPron, LatI, LatAnom, IndEurU, IndEurO, LatI2, IndEurI, LatA, IndEurE, IndEurA, LatX, Ind, IndEurX, LatE & IndEurA,IndEurO, IndEurInd \\\hline
        InflClass[nominal] & IndEurX & IndEurI, IndEurO, Ind, IndEurA & IndEurA,IndEurO, IndEurU\\\hline
        Mood &  & Ind, Sub, Imp &  \\\hline
        NameType & Lit, Ast, Oth, Met, Giv, Nat, Let, Rel, Cal, Com, Sur, Geo &  &  \\\hline
        Number &  & Plur, Sing & Plural \\\hline
        Number[psor] &  & Plur, Sing &  \\\hline
        NumForm & Reference, Word & Roman &  \\\hline
        NumType &  & Card, Dist, Mult, Ord &  \\\hline
        NumValue & 2 &  &  \\\hline
        PartType &  & Int, Emp &  \\\hline
        Person &  & 2, 3, 1 &  \\\hline
        Person[psor] &  & 2, 3, 1 &  \\\hline
        Polarity &  &  & Neg \\\hline
        Poss &  & Yes &  \\\hline
        PronType & Ind, Rel, Art, Rcp & Tot, Neg, Con, Prs, Rel, Int, Dem, Ind & Emp \\\hline
        Proper & Yes &  &  \\\hline
        Reflex &  & Yes &  \\\hline
        Tense &  & Past, Pqp, Fut, Pres &  \\\hline
        Typo & Yes &  &  \\\hline
        UPOS & PUNCT & SCONJ, ADP, ADJ, AUX, VERB, X, NUM, \_, PART, INTJ, ADV, NOUN, DET, CCONJ, PROPN, PRON &  \\\hline
        Variant &  &  & Greek \\\hline
        VerbForm & Conv, Vnoun & Fin, Inf, Part & Ger, Gdv, Sup \\\hline
        VerbType & Mod &  &  \\\hline
        Voice &  & Pass, Act &  \\
        \bottomrule
    \end{tabular}
    \caption{Feature and Values Comparison between UD and LASLA. Note that Perseus and PROIEL (the only UD treebanks that overlap with LASLA) lack some feature values that the other UD treebanks have, but this shows the union of all UD features.}
    \label{tab:feature_comparison_ud_lasla}
\end{table*}

\begin{table*}
\centering
\footnotesize
\setlength{\tabcolsep}{2pt}
\begin{tabular}{|>{\raggedright\arraybackslash}p{4cm}r|>{\raggedright\arraybackslash}p{4cm}r|>{\raggedright\arraybackslash}p{4cm}r|}
\toprule
\multicolumn{2}{|c|}{\textbf{Classical (UD Only)}} & \multicolumn{2}{c|}{\textbf{Bible}} & \multicolumn{2}{c|}{\textbf{Post Classical}} \\
\hline
Work & \# Sents & Work & \# Sents & Work & \# Sents \\
\hline
BellumGallicum & 1445 & jerome\_vulgata-Mark & 1257 & aquinas\_summa-contra-gentiles & 23687 \\
DeOfficiis & 557 & jerome\_vulgata-1-John & 12 & dante\_de-vulgari-eloquentia & 419 \\
InCatilinam & 137 & jerome\_vulgata-2-John & 3 & dante\_letters & 376 \\
Metamorphoseon & 183 & jerome\_vulgata-3-John & 4 & dante\_questio-de-aqua-et-terra & 133 \\
PetroniusSatiricon & 547 & jerome\_vulgata-John & 1765 & dante\_eclogues & 111 \\
PropertiusElegiae & 224 & jerome\_vulgata-Luke & 2044 & llct\_39 & 165 \\
Catilina & 336 & jerome\_vulgata-Galatians & 189 & llct\_79 & 670 \\
TacHistoriae & 64 & jerome\_vulgata-Titus & 39 & palladius\_opus-agriculturae & 955 \\
Aeneis & 68 & jerome\_vulgata-1-Thessalonians & 97 & llct\_36 & 276 \\
cicero\_letters-to-atticus-1 & 703 & jerome\_vulgata-James & 7 & llct\_80 & 571 \\
cicero\_letters-to-atticus-2 & 800 & jerome\_vulgata-Acts & 1490 & llct\_72 & 271 \\
cicero\_letters-to-atticus-4 & 703 & jerome\_vulgata-Hebrews & 13 & llct\_83 & 812 \\
cicero\_letters-to-atticus-5 & 688 & jerome\_vulgata-Colossians & 29 & llct\_73 & 518 \\
cicero\_letters-to-atticus-6 & 270 & jerome\_vulgata-revelation & 763 & llct\_40 & 324 \\
 &  & jerome\_vulgata-1-Corinthians & 736 & llct\_84 & 771 \\
 &  & jerome\_vulgata-2-Peter & 2 & llct\_86 & 826 \\
 &  & jerome\_vulgata-Matthew & 1978 & llct\_75 & 462 \\
 &  & jerome\_vulgata-2-Corinthians & 345 & llct\_38 & 276 \\
 &  &  &  & llct\_74 & 404 \\
 &  &  &  & llct\_81 & 288 \\
 &  &  &  & llct\_77 & 333 \\
 &  &  &  & llct\_76 & 216 \\
 &  &  &  & llct\_85 & 807 \\
\textbf{Total} & 6725 & \textbf{Total} & 10773 & \textbf{Total} & 33671 \\
\hline
phaedrus\_fabulae & 389 & jerome\_vulgata-1-Peter & 5 & aquinas\_forma & 3290 \\
augustus\_res-gestae & 38 & jerome\_vulgata-1-Timothy & 4 & dante\_monarchia & 682 \\
suetonius\_life-of-augustus & 109 & jerome\_vulgata-2-Thessalonians & 37 & llct\_37 & 170 \\
cicero\_letters-to-atticus-3 & 420 & jerome\_vulgata-2-Timothy & 47 & llct\_78 & 389 \\
cicero\_letters-to-atticus-7 & 85 & jerome\_vulgata-Ephesians & 100 & llct\_82 & 472 \\
 &  & jerome\_vulgata-Jude & 22 &  &  \\
 &  & jerome\_vulgata-Philemon & 25 &  &  \\
 &  & jerome\_vulgata-Philippians & 97 &  &  \\
 &  & jerome\_vulgata-Romans & 684 &  &  \\
\textbf{Total} & 1041 & \textbf{Total} & 1021 & \textbf{Total} & 5003 \\
\bottomrule
\end{tabular}

\caption{Number of UD sentences in our custom train (top) and test (bottom) splits. Works that appear only in LASLA are not listed, as there are too many. See \protect\hyperlink{https://www.lasla.uliege.be/cms/c_8570472/fr/lasla-textes-latins-traites-par-auteur}{LASLA's website} for a full list. }
\label{tab:custom_splits_by_work}
\end{table*}

\end{document}